\documentclass[runningheads]{llncs}
\usepackage{graphicx}
\usepackage{booktabs}
\usepackage{amsmath}
\usepackage{amssymb}
\usepackage[dvipsnames]{xcolor}
\usepackage{thmtools}
\usepackage{comment}
\usepackage{soul}
\usepackage{algpseudocode}
\usepackage{algorithm}
\usepackage{bbold}
\usepackage{pdflscape}
\usepackage{bm}
\usepackage{arydshln}
\usepackage{hyperref}
\usepackage{mathtools}

\usepackage{bm}

\usepackage{xspace}
\newcommand{\ie}{\emph{i.e.}\xspace}

\newcommand{\CostRC}{\textsf{CROCO}\xspace}
\newcommand{\cf}[1]{\ensuremath{\breve{#1}}}
\DeclareMathOperator{\E}{\mathbb{E}}
\DeclareMathOperator{\Prob}{\mathbb{P}}

\begin{document}
\title{Generating robust counterfactual explanations}
%
%
\author{
Victor Guyomard\inst{1,2}
\and Françoise Fessant\inst{1}
\and Thomas Guyet\inst{3}
\and Tassadit Bouadi\inst{2}
\and Alexandre Termier\inst{2}
}
%
\authorrunning{Paper N°639}
%
\institute{
Orange Innovation, Lannion, France\\
\email{victor.guyomard@orange.com}\\
\and
Univ Rennes, Inria, CNRS, IRISA, Rennes, France \and
Inria, AIstroSight, France
}
\maketitle              
\begin{abstract}
Counterfactual explanations have become a mainstay of the XAI field. This particularly intuitive statement allows the user to understand what small but necessary changes would have to be made to a given situation in order to change a model prediction. The quality of a counterfactual depends on several criteria: realism, actionability, validity, robustness, etc. In this paper, we are interested in the notion of robustness of a counterfactual. 
More precisely, we focus on robustness to counterfactual input changes. This form of robustness is particularly challenging as it involves a trade-off between the robustness of the counterfactual and the proximity with the example to explain. We propose a new framework, \CostRC, that generates robust counterfactuals while managing effectively this trade-off, and guarantees the user a minimal robustness. An empirical evaluation on tabular datasets confirms the relevance and effectiveness of our approach.
\end{abstract}

\section{Introduction}
The ever-increasing use of machine learning models in critical decision-making contexts, such as health care, hiring processes or credit allocation, makes it essential to provide explanations for the individual decisions made by these models. To this end, Wachter et al. proposed counterfactual explanation~\cite{Wachter2017CounterfactualEW}. A counterfactual is defined as the smallest modification of feature values that changes the prediction of a model to a given output. 
The counterfactual can provide actions (or recourse) for individuals to attain more desirable outcomes. 
This is particularly important in areas where decisions made by algorithms can have significant impacts on people’s lives such as finance, health care or criminal justice. Many methods have been proposed to generate counterfactuals, focusing on some specific properties such as realism ~\cite{PawelczykWWW20,VanLooverenECML21,Guyomard2022VCNetAS}, actionability~\cite{ustun2019actionable,poyiadzi2020face} or sparsity ~\cite{brughmans2021nice,Wachter2017CounterfactualEW,mothilal2020explaining}.  
According to Artelt et al.~\cite{artelt2021evaluating}, many counterfactual generation methods are vulnerable to small changes, where even a minor change in the value of a counterfactual feature can cause the counterfactual to have a different outcome.
Such a situation may arise for example in practical implementation of the counterfactual, due to various factors such as unexpected noise, or adversarial manipulation. As an illustration, a counterfactual may suggest to an individual to raise its salary by 200\$ to obtain a credit, but in practice, the salary is increased by 199\$ or 201\$, potentially resulting in a negative decision (a rejected credit) regarding the decision model. This line of discussions falls into the topic of robustness~\cite{recourse_invalidation,pmlr-v162-dominguez-olmedo22a,VIRGOLIN2023103840,robustness_region}. 
To address robustness in the context of counterfactual explanation, Pawelcyk et al.~\cite{recourse_invalidation} introduce the notion of recourse invalidation rate which represents the probability of obtaining a counterfactual with a different predicted class, when small changes (sampled from a noise distribution) are applied to it. 
They presented an estimator of the recourse invalidation rate in the context of Gaussian distributions, and also a framework (PROBE) that guarantees the recourse invalidation rate to be no greater than a  target specified by the user. 
A limitation of their approach is that the satisfaction of the user condition is dependent of the estimator quality, which means that in practice, the recourse invalidation rate can be greater than the target fixed by the user.
Moreover, PROBE leads in practice to a poor trade-off management between proximity and robustness i.e the counterfactual is robust but far from the example to explain. 
In this paper, we introduce a framework called \CostRC (Cost-efficient RObust COunterfactuals), which is based on a new minimization problem inspired by PROBE~\cite{recourse_invalidation}. 
Our framework introduces the novel concept of soft recourse invalidation rate, as well as an estimator of it.
It enables us to derive an upper-bound for the recourse invalidation rate with almost certain probability. This ensures that the user obtains a solution with a recourse invalidation rate lower than the predetermined target. An experimental evaluation on different tabular datasets confirms these theoretical results, and shows that our method better optimizes the two criteria of robustness and proximity.

\section{Related work} \label{sec:related_work}
Since Wachter et al. seminal paper~\cite{Wachter2017CounterfactualEW}, a variety of counterfactual explanation technics have been proposed.
These methods seek to enhance the quality of counterfactuals by incorporating additional properties, such as constraining the counterfactual to support the data distribution in order to produce realistic examples, freezing immutable features (such as race or gender), producing multiple counterfactuals at once, or even adding causality constraints. We refer the readers to Guidotti et al.~\cite{guidotti2022counterfactual} for a detailed review about counterfactual explanation properties and methods.
The property of robustness has been studied recently in the context of counterfactual explanations, where the validity of a counterfactual is determined by its ability to maintain the same predicted class in the presence of changes. Mishra et al.~\cite{mishra2021survey} distinguish various types of robustness:
\begin{description}
\item[Robustness to model change] refers to the evolution of the validity of the counterfactual explanation when machine learning models are re-trained or when training parameters settings are slightly modified.
Rawal et al.~\cite{rawal2020algorithmic} have demonstrated that state-of-the-art counterfactual generation methods have the tendency to produce solutions that are not robust to model retraining. To address this problem, Ferrario and Loi~\cite{Ferrario2022} proposed to use counterfactual data augmentation every time machine learning models are retrained. Upadhyay et al.~\cite{upadhyay2021towards} for their part developed an adversarial training objective that produces counterfactuals that are robust regarding changes in the training data. More specifically, they evaluated the robustness on different types of training data shift which are data correction shift, temporal shift, and geospatial shift. However, the counterfactuals that are generated suffer from a much higher cost of change regarding state-of-the art counterfactual generation methods~\cite{recourse_invalidation}. In the context of slightly changed training settings, Black et al.~\cite{robustness_deep} achieved robust counterfactual explanations with a regularization method based upon a $K$-Lipschitz constant.
\item[Robustness to input perturbations] refers to how counterfactuals explanations are sensitive to slight input changes. 
According to Dominguez-Olmedo et al.~\cite{pmlr-v162-dominguez-olmedo22a}, a counterfactual is said robust if small changes in the example to explain result in valid counterfactuals. They proposed an optimization problem that applies to linear models and neural networks to generate robust counterfactuals in this context. 
For Artelt et al.~\cite{artelt2021evaluating} robustness means that two examples that are close, must result in two similar counterfactuals. To address this issue they propose to solve an optimization problem that includes a density constraint~\cite{artelt2021evaluating}. They empirically show that having a counterfactual that lies in a dense area has the effect of improving the robustness. Laugel et al.~\cite{laugel2019issues} pointed out that such a type of robustness issue cannot solely be attributed to the explainer, but also arises from the decision boundary of the classifier, thus increasing the problem complexity.

\item[Robustness to counterfactual input changes] refers to the ability of a counterfactual explanation to remain valid when small feature changes are applied (two similar counterfactuals should have the same predicted class). In this context, Pawelcyk et al.~\cite{recourse_invalidation} presented PROBE a framework to produce robust counterfactuals that is based on an optimization problem. This framework aims to find a trade-off between two criteria that are the recourse invalidation rate and the proximity, i.e. the distance between the counterfactual and the example to explain. 
From their side, Maragno et al.~\cite{robustness_region} introduced an adversarial robust approach that generates counterfactuals that remain valid in an uncertainty set, meaning that for a given example to explain, all the solutions in the set are valid counterfactuals. This approach works for non-differentiable model unlike PROBE. However there is no trade-off between the recourse invalidation rate and the proximity as all the counterfactuals in the uncertainty set are valid.  In such a scenario, the robustness constraint cannot be relaxed, then allowing the generation of counterfactuals that are far from the example to explain. 
Our approach, \CostRC, is part of this category of methods. It is inspired by the PROBE framework, and improves its limitations. Indeed, the major criticism that we can make to PROBE is that the guarantees in terms of robustness that it offers to the user are completely dependent on the quality of their estimator (i.e. the guarantee is based
on a recourse invalidation rate approximation rather than the true recourse invalidation rate). Our method introduces a new optimization problem that is proved to induce an almost-sure upper bound on the true recourse invalidation rate. This leads to a significant improvement in the trade-off between the robustness of the counterfactual and the proximity with the example to explain.

\end{description}
\section{Problem statement}

In this section, we define some notations related to the generation of counterfactuals, and we formalize the robustness of counterfactual generation by introducing the notion of \textit{recourse invalidation rate}. 

\subsection{Generation of counterfactuals}

We consider the generation of counterfactuals for a binary classifier. 
Let $\mathcal{X}\subseteq \mathbb{R}^n$ represents the $n$-dimensional feature space. 
A binary classifier is a function $h: \mathcal{X} \rightarrow \mathcal{Y}$ where $\mathcal{Y}=\{0,1\}$. 
We assume that the classification is obtained from a probabilistic prediction i.e. a function $f: \mathcal{X} \rightarrow [0,1]$ that returns $\hat{p}$ which is the predicted probability for the class~$1$. 
Then, the predicted class is the most likely class according to $\hat{p}$. For a given example $x$, $h({x})=g \circ f({x})$ where $g: [0,1] \rightarrow \mathcal{Y}$ is a function that returns the predicted class from the probability vector. 
We take $g(u)=\mathbb{1}_{>t}(u)$, where $t$ is the decision threshold.  $\mathbb{1}_{>t}(u)$ equals $1$ if $u>t$ and $0$ otherwise.

In this article, we do post-hoc counterfactual generation, meaning that $f$ (and thus $h$) are given. 
And for a given example to explain $x\in\mathcal{X}$, whose decision is $h(x)$, we want to generate a counterfactual $\cf{x}\in \mathcal{X}$. A counterfactual is a new example close to the example to explain $x$, and with a different prediction, \ie $h(\cf{x})\neq h(x)$. 
If it is true that $h\left(\cf{x}\right)\neq h\left (x\right)$, then $\cf{x}$ is said to be \textit{valid}.
A counterfactual $\cf{x}$ is also seen as a change to apply to $x$: $\cf{x}=x+\delta$ where $\delta\in\mathbb{R}^n$. 
Thus, a counterfactual is associated to a small change $\delta$ that modifies the decision returned by $h$. 
Generating a counterfactual is basically solving the following optimisation problem:
\begin{equation} \label{eq:counterfactual_generation}
    \min_{\delta} \ell\left(f\left(x+\delta\right), 1-h(x)\right)+\lambda \left\|\delta\right\|_1
\end{equation}
 where $\ell:[0,1]^2\mapsto\mathbb{R}^+$ quantifies the distance between the predicted probability, $f\left(\cf{x}\right)$, and $1-h(x)$ that is the opposite of the predicted class for example $x$. 
 For instance, Wachter et al. suggested  $\ell$ as the $L_2$ distance, so as to produce counterfactuals that are close to the desired decision~\cite{Wachter2017CounterfactualEW}. The other term in the optimization problem, constraints the change $\delta$ applied to the example $x$ to be small.

In what follows, we will focus specifically on the generation of counterfactuals in the case of instances that have received a negative decision (which corresponds to instances predicted as class $0$). This choice has no limitation and is motivated by the fact that the majority of robustness methods are defined in a recourse context~\cite{recourse_invalidation,rawal2020algorithmic,upadhyay2021towards} where the goal is to provide explanations only for negatively predicted instances. 
We will also assume that the classifier $f$ is differentiable.

\subsection{Recourse invalidation rate}

In order to quantify the robustness of the counterfactual to an input perturbation, the notion of recourse invalidation rate has been introduced by Pawelczyk et al.~\cite{recourse_invalidation}.
\begin{definition}[Recourse invalidation rate] \label{def:recourse_invalid_rate}
The recourse invalidation rate for a counterfactual $\cf{x}$, of an example $x$ predicted as class $0$ can be expressed as:
\begin{equation*}
    \Gamma\left(\cf{x};p_{\varepsilon}\right)=\E_{\varepsilon\sim p_\epsilon}\left[1-h\left(\cf{x}+\varepsilon\right)\right]
\end{equation*}
where $\varepsilon\in \mathbb{R}^{n}$ is a random variable that follows a probability distribution $p_{\varepsilon}$. Since $h\left(\cf{x}+\varepsilon\right) \in \left\{ 0,1\right\} $, it ensues $\Gamma(\cf{x};p_{\varepsilon}) \in [0,1]$.
\end{definition}

Assuming $p_\varepsilon$ is centered, then $p_\varepsilon$ defines a region around a counterfactual $\cf{x}$ for \textit{similar} counterfactuals $\cf{x}+\varepsilon$. 
Intuitively, $\Gamma(\cf{x};p_{\varepsilon})$ gives the rate of \textit{similar} counterfactuals that are not valid, \ie that belong to class~$0$. 
Thus, the lower $\Gamma(\cf{x};p_{\varepsilon})$, the more robust is the counterfactual. 
If $\Gamma\left(\cf{x};p_{\varepsilon}\right) = 0$, the counterfactual is considered perfectly robust, given that all the perturbed counterfactuals result in positive outcomes (i.e., there are all predicted as class~$1$).
However, if $\Gamma\left(\cf{x};p_{\varepsilon}\right) = 1$, the counterfactual is not at all considered robust, since no noisy counterfactuals lead to positive outcomes (i.e., there are all predicted as class~$0$).

\begin{figure}[t!]
    \centering
    \includegraphics[width=\textwidth]{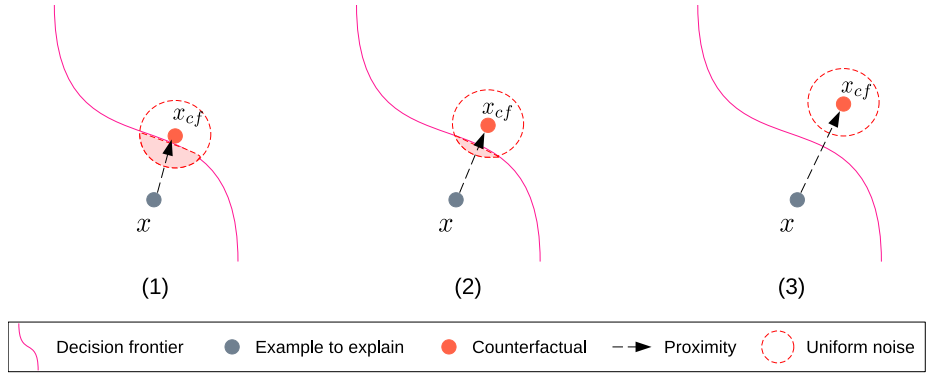}
    \caption{Illustration of the recourse invalidation rate with a uniform distribution $p_\varepsilon$ (dashed-red circle). The recourse invalidation rate is figured out by the area of the region in red. 
    In \textbf{(1)} the counterfactual has a low robustness and is at a low distance from the example. In \textbf{(2)} the counterfactual has a medium robustness and is at a medium distance, and in \textbf{(3)} the counterfactual has a perfect robustness but is far from the example (large distance).}
    \label{fig:trade_off}
\end{figure}

Figure~\ref{fig:trade_off} illustrates the intuition of the recourse invalidation rate. 
$\Gamma(\cf{x};p_{\varepsilon})$ can be seen as the surface of the neighborhood that overlaps the region, split by the decision frontier, on the side of the example. This neighborhood represents the perturbations on the counterfactuals that we would like to accept without changing its validity. 
The Figure also shows that finding a robust counterfactual requires to make a trade-off between the robustness and the magnitude of the change.

\subsection{The PROBE framework for generating robust counterfactuals} 
Pawelczyk et al.\cite{recourse_invalidation} have developed a framework named PROBE that generates robust counterfactuals regarding the recourse invalidation rate. 
It adapts the minimization problem of equation~\ref{eq:counterfactual_generation} by adding a new term that enforces the recourse invalidation rate to be under a target value $\Gamma_{t}$. 
This target value is chosen by the user. 
More formally, generating a counterfactual relies on solving the following minimization problem:
\begin{equation} \label{eq:PROB_problem}
\min_{\delta}\;\; \max\left[\Gamma\left(x+\delta;p_{\varepsilon}\right) - \Gamma_t,\,0\right]+\ell\left(f\left(x+\delta\right),\,1-h(x)\right)+\lambda \left\|\delta\right\|_1
\end{equation}
There are some difficulties with the additional constraint on recourse invalidation rate. Indeed, the true value of $\Gamma$ can not be evaluated in practice. Then, PROBE proposes a Monte-Carlo estimator of $\Gamma$. This means that it is estimated by computing the mean of a sample of perturbations in $p_{\varepsilon}$: 
\begin{equation} \label{eq:monte_carlo_estimator}
 \tilde{\Gamma}\left(\cf{x};K,p_\varepsilon\right) = \frac{1}{K} \sum_{k=1}^K \left(1 - h\left(\cf{x} + \varepsilon_k\right)\right)
\end{equation}

However, $\tilde{\Gamma}$ is non-differentiable, because $h({x})=g \circ f({x})$ and $g(u)=\mathbb{1}_{>t}$. Then, it can not be part of a loss of an optimization problem. 
To overcome this limitation, the authors proposed a first-order approximation of the true recourse invalidation rate $\Gamma$ in the context of a Gaussian distribution noise $p_\varepsilon = \mathcal{N}(\mathbf{0}, \sigma \mathbf{I})$, named $\tilde{\Gamma}_{\text{PROBE}}$. 

\begin{figure}[t!]
    \centering
    \includegraphics[width=0.2\textwidth]{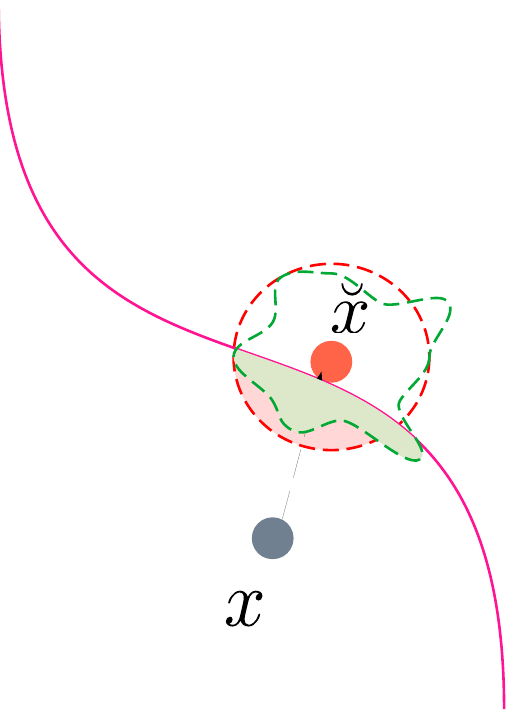}
    \caption{Illustration of the potential problem with PROBE. The red region illustrates the true recourse invalidation rate (see Figure~\ref{fig:trade_off}) while the green region illustrates the approximated recourse invalidation rate through the approximation of the red region. In this case, the approximation under-estimates the red region and misleadingly encourages finding a $\cf{x}$ that would break the robustness constraint.}
    \label{fig:PROB_test}
\end{figure}

Then, the optimization algorithm solves the problem in eq.~\ref{eq:PROB_problem}, replacing $\Gamma$ by $\tilde{\Gamma}_{\text{PROBE}}$ and stops when the approximation of recourse invalidation rate is under the target value, i.e. when $\tilde{\Gamma}_{\text{PROBE}}(\cf{x;p_\varepsilon}) \leq \Gamma_{t}$.

Thus, for a given counterfactual $\cf{x}$ returned by PROBE, the user is guaranteed that $\tilde{\Gamma}_{\text{PROBE}}(\cf{x};p_\varepsilon) \leq \Gamma_{t}$. However, this means that the guarantee depends on the quality of the estimator. 
Indeed, it is possible to generate a counterfactual where 
$\tilde{\Gamma}_{\text{PROBE}}(\cf{x};p_\varepsilon) \leq \Gamma_{t} \leq \Gamma(\cf{x};p_\varepsilon)$ which would then violate the user-selected guarantee. The intuition behind this situation is depicted in Figure~\ref{fig:PROB_test}.

To sum up, PROBE has two limitations: 1) It offers users a guarantee based on the recourse invalidation rate approximation rather than the true recourse invalidation rate; 2) the approximation applies only for Gaussian distribution of counterfactual perturbation. This makes the approach difficult to extend to categorical attributes. 

Our contribution overcomes the first limitation by introducing a new estimator that is proved to induce an almost-sure upper bound on the true recourse invalidation rate. 
Furthermore, our approach is independent to the noise distribution, thus enabling the use of various noise distributions.

\section{Our contribution}
In this section, we present our method, named \CostRC standing for \textit{Cost-efficient RObust COunterfactuals}. 
It improves the generation of robust counterfactuals according to the recourse invalidation rate. 

This method, inspired from PROBE, introduces a new robustness term to the optimization problem presented in Equation~\ref{eq:counterfactual_generation}. 
This term is based on an upper-bound of the recourse invalidation rate.

\subsection{An upper bound of the recourse invalidation rate}
As it is not feasible to derive a closed-form expression of $\Gamma$ without making any assumption about the noise distribution, and given that $\tilde{\Gamma}$ is not differentiable, our idea is to compute an upper-bound of $\Gamma$.

Let $\cf{x}$ be a counterfactual for an example $x\in\mathcal{X}$, then we define the soft recourse invalidation rate, $\Theta(\cf{x})$ by:
$$\Theta(\cf{x};p_{\varepsilon})= \E_{\varepsilon\sim p_{\varepsilon}}\left[1-f\left({\cf{x}}+\varepsilon\right)\right].$$
The proposition~\ref{res:upper_lower_our} states that the soft recourse invalidation rate, $\Theta$, induces an upper-bound of the recourse invalidation rate, $\Gamma$. 

\begin{proposition}\footnote{All proofs are provided in Section A.1 of supplementary material.} \label{res:upper_lower_our}%
Let $t\in[0,1]$ be a decision threshold and $\cf{x}$ be a counterfactual for an example $x\in\mathcal{X}$, an upper bound of the true recourse invalidation rate is given by:
\begin{equation}\label{eq:upper_lower_our}%
\Gamma\left(\cf{x};p_{\varepsilon}\right)\leq \frac{\Theta\left(\cf{x};p_{\varepsilon}\right) }{\left( 1-t\right) }
\end{equation}
\end{proposition}

Similarly to $\Gamma$, $\Theta$ can not be evaluated directly. 
However, we can use the following Monte-Carlo estimator, where $K$ is the number of random samples:
\begin{equation} \label{eq:our_estimator} 
 \tilde{\Theta}\left(\cf{x};K,p_{\varepsilon}\right) = \frac{1}{K} \sum_{k=1}^K (1 - f(\cf{x} + \varepsilon_k))
\end{equation}
This quantity can be seen as the mean predicted probability for class~$0$, computed on perturbed samples that  are randomly drawn from the $p_{\epsilon}$ distribution. 
The proposed estimator is close to the recourse invalidation rate estimation outlined in equation~\ref{eq:monte_carlo_estimator}, but it differs in that it is differentiable as a composition of differentiable functions, thus can be included in an objective function.

Moreover, the proposition~\ref{ref:our_estimator_quality} shows that our estimator, $\tilde{\Theta}$, defines an almost-sure upper bound of the true recourse invalidation rate. 
This means that $\frac{m+\tilde{\Theta}}{1-t}$ has a high probability to be an upper-bound of $\Gamma$. 

\begin{proposition}\label{ref:our_estimator_quality}
Let $t\in[0,1]$ be a decision threshold, $p_{\varepsilon}$ a noise distribution, $\cf{x}$ be a counterfactual for an example $x\in\mathcal{X}$, then an almost-sure upper-bound of the recourse invalidation rate is given by:
\begin{equation}
\mathbb{P}\left( \Gamma\left(\cf{x};p_{\varepsilon}\right) \leq \dfrac{m+\tilde{\Theta}\left(\cf{x};K,p_{\varepsilon}\right)}{1-t}\right) \geq 1-\exp \left( -2m^{2}K\right) 
\end{equation}
where $m>0$ and $K$ is the number of random samples.
\end{proposition}

With a high number of random samples and a given value of $m$, the exponential term of proposition~\ref{ref:our_estimator_quality} can be arbitrarily small.
Then for a given value of our estimator $\tilde{\Theta}\left(\cf{x};K,p_{\varepsilon}\right)$, we have almost surely that the true recourse invalidation rate will be in the worst case equals to $\dfrac{m+\tilde{\Theta}\left(\cf{x};K,p_{\varepsilon}\right)}{1-t}$. 
It ensues that if we enforce $\dfrac{m+\tilde{\Theta}\left(\cf{x};K,p_{\varepsilon}\right)}{1-t}$ to be lower  than a given threshold $\bar\Gamma_{t}$, then we are almost-sure that the true recourse invalidation rate is lower than $\bar\Gamma_{t}$, \ie that the counterfactual is more robust than the given threshold.

Note that $m\in\mathbb{R}_{>0}$ is a parameter that defines the tightness of the upper-bound. The lower $m$, the better the upper-bound. In return, low $m$ requires a higher $K$ (\ie more computational resource) to keep the confidence in the bound. Section A.2 in supplementary material provides a table to choose the values of $m$ and $K$ with respect to the desired level of confidence.

For instance, with $K=500$ and $m=0.1$, and $t=0.5$, the inequation of the proposition~\ref{ref:our_estimator_quality} gives:
\begin{equation}\label{eq:example_of_guarantee}
\Prob\left( \Gamma\left(\cf{x}\right) \leq 0.2+2\tilde{\Theta}\left(\cf{x}\right) \right) \geq 0.999 
\end{equation}

\subsection{Generate robust counterfactuals} \label{sec:our_opt_problem}
We propose a minimization problem for the generation of robust counterfactuals according to the recourse invalidation rate. 

Given a neighborhood distribution $p_{\varepsilon}$, a number of samples $K$, a tightness value $m>0$ and a target upper-bound $\bar\Gamma_t$, a counterfactual $\cf{x}=x + \delta$ is found by minimizing the following objective function:
\begin{equation} \label{eq:optimization_problem}
    \min_{\delta} \underbrace{\left(\frac{\tilde{\Theta}\left(x+\delta;K,p_{\varepsilon}\right)+m}{1-t}-\bar\Gamma_{t}\right)^{2}}_{\text{Robustness}} +\underbrace{\ell\left(f\left(x+\delta\right), 1-h(x)\right)}_\text{Validity} +\underbrace{\lambda \left\|\delta\right\|_1}_\text{Proximity}
\end{equation}

The last two terms implement the classical trade-off for counterfactual generation. Indeed, the second term pushes the counterfactual class toward a class that differs from the example class (if $h(x)=0$ then we want $h(\cf{x})=1$), while the last term minimizes the distance between the counterfactual and the example to explain. 

The first term encourages our new estimator to be close to a target value $\bar\Gamma_{t}$, \ie the target upper-bound of the recourse invalidation rate. This pushes to choose a counterfactual that has an upper bound close to the objective.

Algorithm~\ref{algorithm:CostRC} describes the optimization process for \CostRC. 
Gradient steps are performed until the counterfactual predicted class is flipped ($f\left(x+\delta\right)\ge t $), and the value of the upper-bound $\frac{m+\tilde{\Theta}\left(x+\delta;K,p_\varepsilon\right)}{1-t}$  is below the target value $\bar\Gamma_{t}$.

\begin{algorithm}[t]
  \caption{\texttt{\CostRC} optimization for counterfactual generation}
  \label{algorithm:CostRC}
\begin{algorithmic}
   \State {\bfseries Input:} $x$ s.t.\ $f(x) < 0$, $f$, $\lambda > 0$, $\alpha$, $\bar\Gamma_{t}>0$, $K$,$p_\varepsilon$
   \State {\bfseries Output:} $x+\delta$
   \State $\delta\gets 0$;
   \State Compute $\tilde{\Theta}\left(x+\delta;K,p_\varepsilon\right)$
  \While{ $f(x+\delta)< 0$ {\bfseries and} $\frac{m+\tilde{\Theta}\left(x+\delta;K,p_\varepsilon\right)}{1-t}> \bar\Gamma_{t}$}
   \State $\delta \gets  \delta - \alpha \cdot \nabla_{\delta} \mathcal{L}_{\CostRC}(x+\delta; \Theta_{t}, p_\varepsilon, \lambda)$ \Comment{From equation \ref{eq:optimization_problem}}
   \State Update $\tilde{\Theta}\left(x+\delta;K,p_\varepsilon\right)$
  \EndWhile
\State {\bfseries Return:} $x+\delta$ 
\end{algorithmic}
\end{algorithm}

\CostRC have several benefits, it allows the user to generate counterfactuals with almost surely a minimal robustness, and this without a hypothesis about the noise distribution. Moreover, our optimization problem relies on an almost-sure upper bound of the true recourse invalidation rate instead of relying on an approximation as Pawelcyk et al. did with PROBE~\cite{recourse_invalidation}. 
Our intuition is that this will in practice improve the trade-off between proximity and robustness.

\section{Experiments and results} \label{sec:exp}

We have divided our experiments into two sections. 
After experimentally confirming that our approach preserves the validity of the counterfactuals, the purpose of the first section is to demonstrate empirically that \CostRC provides an effective management of the trade-off between proximity and robustness in comparison to PROBE.\footnote{\url{https://github.com/twi09/CROCO}}
In the second section, we demonstrate experimentally that the counterfactuals returned by \CostRC exhibits a lower degree of invalidation with respect to the user-defined target than PROBE do. 

First of all, we describe the datasets that we used for evaluation, along with the metrics we employed as well as the predictive model details. 

\subsection{Experimental setting}
For a fair comparison, we used the CARLA library~\cite{pawelczyk2021carla}, which was also used for evaluating PROBE. It contains three binary classification datasets: \textit{Adult}, \textit{Give Me Some Credit} (GSC), and \textit{COMPAS}. These datasets contain both numerical and categorical features. Both numerical and categorical variables are used to train the classifier, but the counterfactuals are generated by modifying only the numerical variables. The proportion of categorical variables for each dataset are respectively 3/7, 1/12 and 25/40.
Additional details about these datasets are available in the section A.4 of the supplementary material.
For every dataset, the classification model $f$, is a fully connected neural network with 50 hidden layers and ReLU activation functions.\footnote{Function carla.models.catalog.MLModelCatalog of the CARLA library.}

We used for evaluation the following metrics:
\begin{description}
\item[Validity] 
    A counterfactual $\cf{x}$ of an example $x$ is valid if the classification model predicts different classes for $x$ and $\cf{x}$~\cite{mothilal2020explaining,mazzine2021framework}. Formally:
    \[
      \text{Validity}=\left\{\begin{array}{l}
      0, \text { if } f(\cf{x})=f(x) \\
      1, \text { if } f(\cf{x}) \neq f(x)
      \end{array}\right.
    \]
    The validity measure lies in $[0,1]$. The higher it is, the better.
\item[Distance]  The distance is the $L_{1}$ distance between an example, $x$ and its counterfactual, $\cf{x}$~\cite{mothilal2020explaining,Wachter2017CounterfactualEW}. 
    \[
        \text{Distance} = \left\|\cf{x}-x\right\|_{1} = \| {\delta} \|_{1}
    \]
   A low value indicates fewer changes of features to apply to the original example to obtain the counterfactual. As the distance decreases, the proximity increases. 
   In the context of counterfactual generation, we assume that the lower the distance, the more actionable the counterfactual, the better.

\item[Recourse invalidation rate] We used $\tilde{\Gamma}$ (see equation \ref{eq:monte_carlo_estimator}) to evaluate recourse invalidation rate, i.e. the robustness of the counterfactual. This value indicates the risk to have an invalid counterfactual in case the counterfactual is slightly changing wrt to the automatically recommended counterfactual. 
The lower, the better. 

The recourse invalidation rate makes the assumption of a neighborhood represented by a distribution, $p_\varepsilon$. \CostRC makes no hypothesis on this distribution but PROBE requires a Gaussian distribution. For the sake of fairness, we use a centered Gaussian distribution with a parameterized variance $\sigma$ for the two methods.
\end{description}

For each dataset, we run PROBE with $\sigma ^{2}\in \left\{0.005,0.01,0.015,0.02\right\}$ and $\Gamma_{t} \in \left\{0.05,0.10,0.15,0.2,0.25,0.3,0.35\right\}$. 
Regarding the setting of \CostRC, we choose $K=500$, $m=0.1$, $t=0.5$. $\lambda$ is found through an iterative procedure that is described in section A.5.2 of supplementary material. 
For each dataset, we run \CostRC with the same parameters as PROBE: $\sigma ^{2}\in \left\{0.005,0.01,0.015,0.02\right\}$ and $\bar\Gamma_{t}\in \left\{0.05,0.10,0.15,0.2,0.25,0.3,0.35\right\}$. 

We also include the approach of Wachter et \textit{al.}~\cite{Wachter2017CounterfactualEW} (referred to as \textbf{Wachter}) in our experiment. This counterfactual generation method establishes a baseline for recourse invalidation rate. 

In our experiments, we generate $500$ counterfactuals for each dataset and each parameterized method. We collected their recourse invalidation rate, distance and validity, that are discussed in the following.

\begin{figure}[t!]
    \centering
    \includegraphics[width=\textwidth,trim={0 7pt 0 20pt},clip]{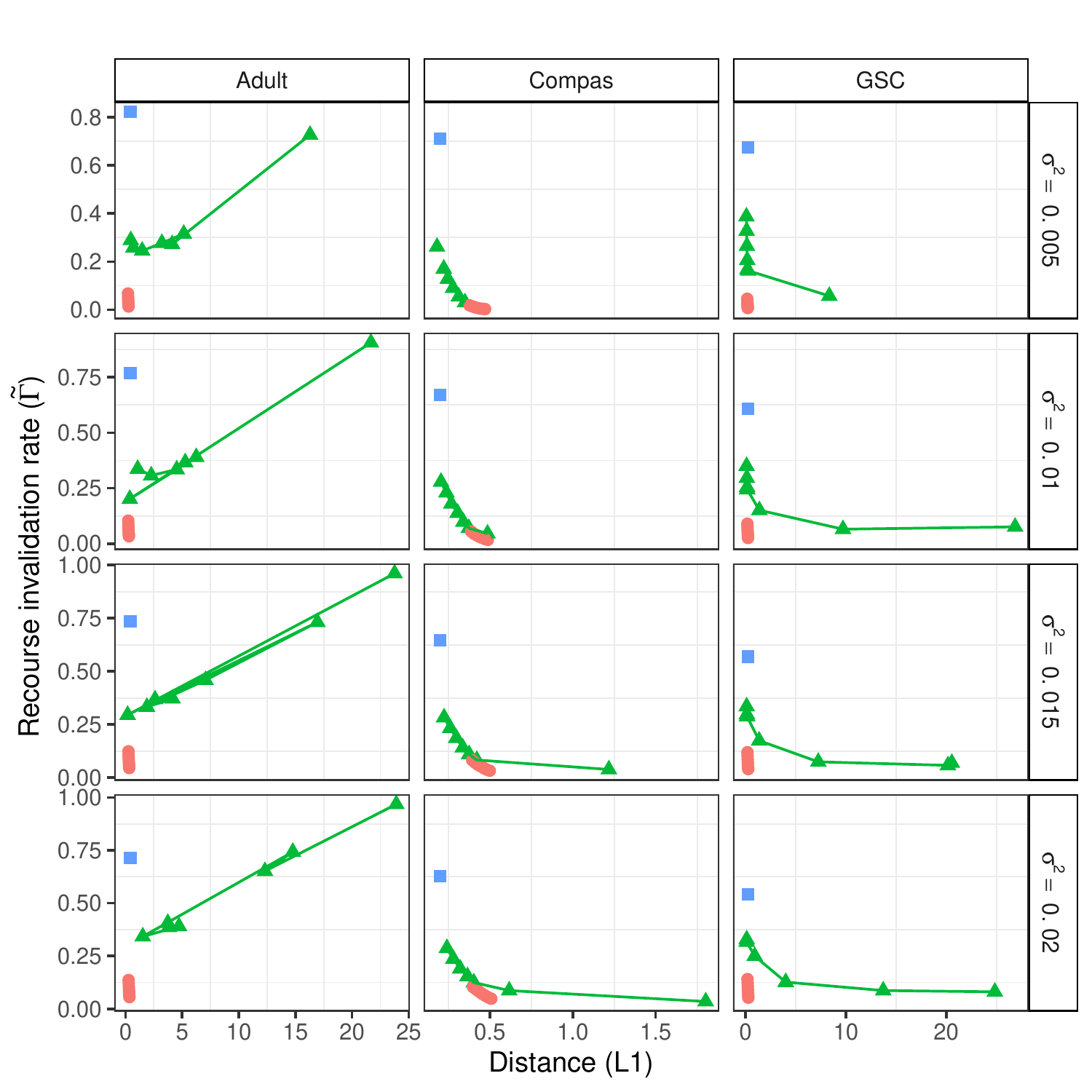}\\
    \includegraphics[width=\textwidth,trim={0 18.5pt 0 22pt},clip]{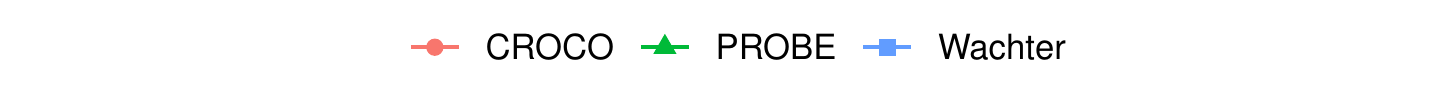}
    \caption{Trade-off between recourse invalidation rate and distance with Gaussian distribution noises. Each column corresponds to a dataset and each line to a value of $\sigma^2 \in \left\{0.005,0.01,0.015,0.02\right\}$. In each subplot the value of  $\sigma ^{2}$ is fixed. Each point of a curve corresponds to a mean recourse invalidation rate and a mean distance for a given target, we have $target\in \left\{0.05,0.10,0.15,0.2,0.25,0.3,0.35\right\}$. The points are connected by target order.}
    \label{fig:trade_off_cost_IR}
\end{figure}

\subsection{Comparisons between PROBE and \CostRC}
In this section, the quality of the counterfactuals generated using  \CostRC, PROBE and Watcher is compared. 

First of all, Watcher and \CostRC achieves a perfect validity for all datasets.
PROBE achieved a perfect validity on all datasets, except for two counterfactual sets, that corresponds to the COMPAS dataset where $\sigma^{2}=0.005$ and $\Gamma_{t}=0.3$ and also the GSC dataset where $\sigma^{2}=0.02$ and $\Gamma_{t}=0.05$.  
As a consequence, in the following, we focus the analysis on the trade-off between the distance and the recourse invalidation rate. The section A.3.1 of the supplementary material contains details regarding the validity obtained for each dataset, and counterfactual sets that are generated. 
 
Figure \ref{fig:trade_off_cost_IR} compares, \textbf{Watcher}, PROBE and \CostRC regarding the distance and recourse invalidation rate on the three different datasets.
Each point of a given curve corresponds to the mean recourse invalidation rate and the mean distance that is obtained from \CostRC or PROBE by fixing a target value. Note that \textbf{Watcher} has only one point as it has no recourse invalidation rate target parameter.
The standard-deviation values are provided in section A.3.2 of  supplementary material.
Note that for a given curve, the points are linked by order of increasing target value. 

For the GSC dataset, \CostRC achieves both smaller distances (higher proximities) and lower recourse invalidation rates compared to PROBE, regardless of the value of  $\sigma ^{2}$.
The same conclusion can be drawn for the COMPAS dataset, except for  $\sigma ^{2}=0.005$ where \CostRC achieves smaller recourse invalidation rates but at the cost of higher distances. 

Regarding the Adult dataset, we observe that PROBE is unstable, as it can produce solutions with higher recourse invalidation rate than the target fixed by the user (where $\Tilde{\Gamma} \geq \Gamma_{t}$). 
Our hypothesis is that the proportion of categorical variables for this dataset makes the generation of counterfactuals difficult based on the numerical variables only.
On the other hand, \CostRC is stable and achieves both smaller distances (higher proximities) and lower recourse invalidation rates. We also noticed that on all the datasets, distance values increase when $\sigma^{2}$ increased, thus confirming the presence of a trade-off between the two quantities. 

When solutions are closely clustered together in terms of mean distances, both PROB and \CostRC exhibit similar standard deviation values. However, when solutions are more widely dispersed, PROB tends to have higher standard deviation values compared to \CostRC (see section A.3.2 of supplementary material).

We observed that for all datasets and values of $\sigma^{2}$, PROBE and \CostRC outperform \textbf{Wachter} in terms of recourse invalidation rates. The only exception is the Adult dataset when $\Gamma_t=0.35$, where PROBE produces higher recourse invalidation rates due to instability issues.

\begin{figure}[tp]
    \centering
    \includegraphics[width=\textwidth,trim={0 6pt 0 20pt},clip]{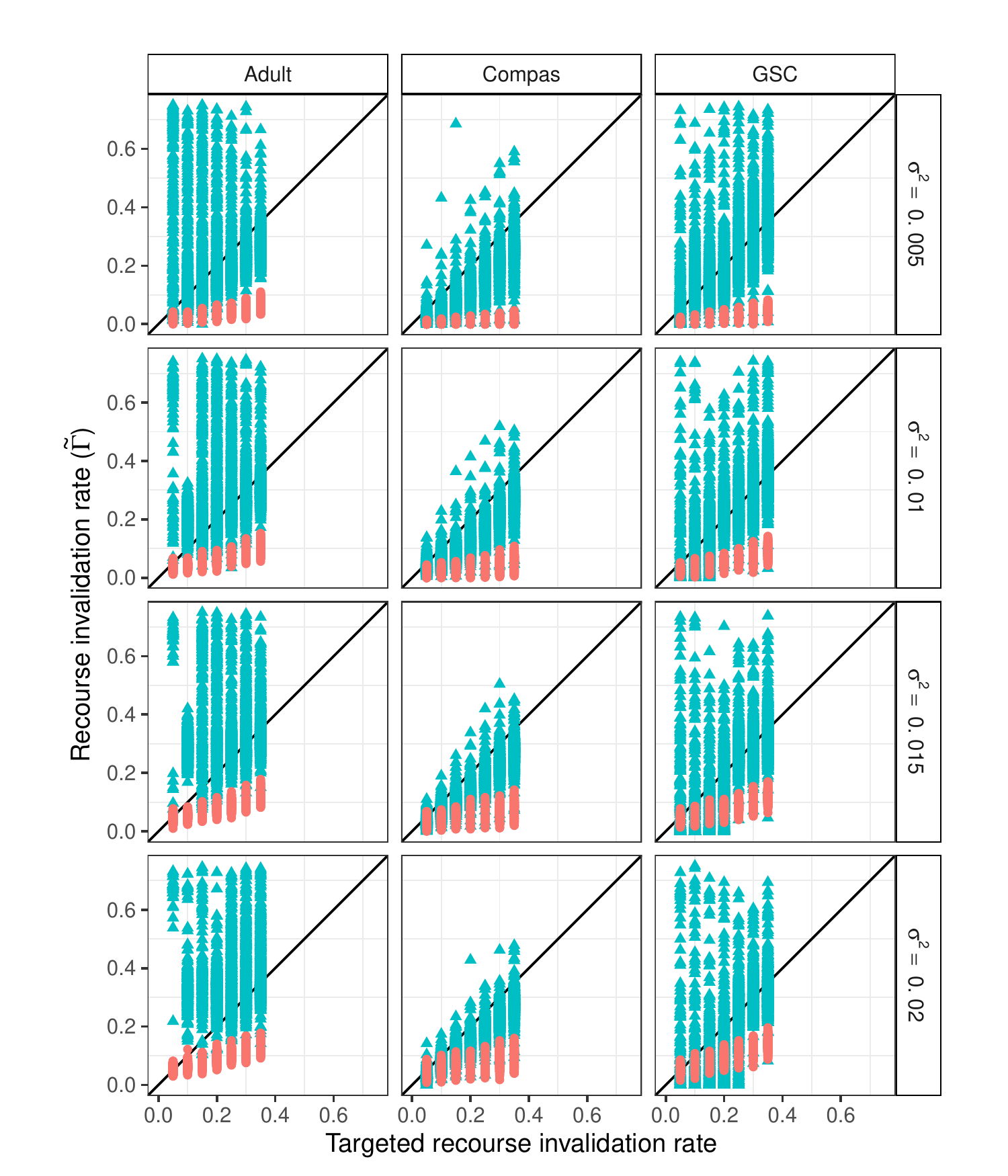}\\
    \includegraphics[width=\textwidth,trim={0 18.5pt 0 22pt},clip]{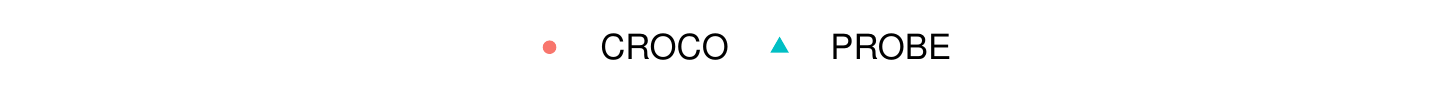}
    \caption{Comparison between targeted recourse invalidation rate and recourse invalidation rate. Each column corresponds to a dataset and each line to a value of $\sigma^2 \in \left\{0.005,0.01,0.015,0.02\right\}$. In each subplot, the value of  $\sigma ^{2}$ is fixed. Each point corresponds to a counterfactual, on the x-axis is presented the target recourse invalidation rate for the counterfactual, and on the y-axis the recourse invalidation rate that is computed.} 
    \label{fig:target_comparisons}
\end{figure}

\subsection{Target invalidation study}
For each counterfactual that is obtained from PROBE or \CostRC, we computed the recourse invalidation rate  and compared it with the targeted recourse invalidation rate.\footnote{Watcher is not figured out as it does not set a target for recourse invalidation rate.} 
The results are provided in Figure~\ref{fig:target_comparisons}. The graphics figure out the diagonal representing the exact match between the targeted and the recourse invalidation rate. All points that are above this diagonal correspond to  counterfactuals that do not achieve the robustness requested by the user. 
We notice that with PROBE, the recourse invalidation rates frequently exceed the target fixed by the user. It illustrates that the approximation of $\Gamma$ made by PROBE is too loose. 
In contrast, for \CostRC, the recourse invalidation rates are typically lower, indicating that the user-specified target is less invalidated.

We computed the upper bound value derived in proposition \ref{ref:our_estimator_quality} for each counterfactual obtained from \CostRC. 

Figure 5 of section A.3.3 of the supplementary material illustrates the evolution of the upper bound value ($\frac{m+\tilde{\Theta}}{1-t}$) with regard to the recourse invalidation rate for different values of $\sigma^2$. Our analysis show that the theoretical bound is not violated. This means that even in cases where \CostRC failed to found a solution that match the user target (i.e. where $\frac{m+\tilde{\Theta}}{1-t}> \bar\Gamma_{t}$), we can still provide the user a guarantee on the true recourse invalidation rate. This guarantee is based on the value of $\tilde{\Theta}$ that is obtained at the end of the optimization.

\section{Conclusion}
In this paper, we introduce \CostRC, a novel framework for generating counterfactuals that are robust to input changes. 
A robust method guarantees that the slightly perturbed counterfactual is still valid.
Our approach leverages a new estimator that 
provides a theoretical guarantee on the true recourse invalidation rate of the generated counterfactuals. Through experiments comparing \CostRC to the state-of-the-art PROBE method, we demonstrate that our approach achieves a better trade-off between recourse invalidation rate and proximity, while also leading to less invalidations regarding the user-specified target.
Moving forward, we plan to extend the capabilities of \CostRC by adapting it to handle categorical variables. Since our approach is independent to the noise distribution, it seems reasonably possible to generate robust counterfactuals for data with both numerical and categorical variables.
\CostRC is implemented in the CARLA framework and will be soon available for practical usage.

%
%
\clearpage
\bibliographystyle{splncs04}

\begin{thebibliography}{10}
\providecommand{\url}[1]{\texttt{#1}}
\providecommand{\urlprefix}{URL }
\providecommand{\doi}[1]{https://doi.org/#1}

\bibitem{artelt2021evaluating}
Artelt, A., Vaquet, V., Velioglu, R., Hinder, F., Brinkrolf, J., Schilling, M.,
  Hammer, B.: Evaluating robustness of counterfactual explanations. In:
  Proceedings of the Symposium Series on Computational Intelligence (SSCI). pp.
  01--09. IEEE (2021)

\bibitem{robustness_deep}
Black, E., Wang, Z., Fredrikson, M.: Consistent counterfactuals for deep
  models. In: Proceedings of the International Conference on Learning
  Representations (ICLR). OpenReview.net (2022)

\bibitem{brughmans2021nice}
Brughmans, D., Leyman, P., Martens, D.: Nice: an algorithm for nearest instance
  counterfactual explanations. arXiv  \textbf{v2} (2021),
  \url{https://arxiv.org/abs/2104.07411}

\bibitem{pmlr-v162-dominguez-olmedo22a}
Dominguez-Olmedo, R., Karimi, A.H., Sch{\"o}lkopf, B.: On the adversarial
  robustness of causal algorithmic recourse. In: Proceedings of the 39th
  International Conference on Machine Learning (ICML). vol.~162, pp. 5324--5342
  (2022)

\bibitem{Ferrario2022}
Ferrario, A., Loi, M.: The robustness of counterfactual explanations over time.
  Access  \textbf{10},  82736--82750 (2022)

\bibitem{guidotti2022counterfactual}
Guidotti, R.: Counterfactual explanations and how to find them: literature
  review and benchmarking. Data Mining and Knowledge Discovery pp. 1--55 (2022)

\bibitem{Guyomard2022VCNetAS}
Guyomard, V., Fessant, F., Guyet, T.: {VCNet}: A self-explaining model for
  realistic counterfactual generation. In: Proceedings of the European
  Conference on Machine Learning and Principles and Practice of Knowledge
  Discovery in Databases (ECML/PKDD). pp. 437--453 (2022)

\bibitem{laugel2019issues}
Laugel, T., Lesot, M.J., Marsala, C., Detyniecki, M.: Issues with post-hoc
  counterfactual explanations: a discussion. arXiv  (2019),
  \url{https://arxiv.org/abs/1906.04774}

\bibitem{robustness_region}
Maragno, D., Kurtz, J., R\"ober, T.E., Goedhart, R., Birbil, S.I., Hertog,
  D.d.: Finding regions of counterfactual explanations via robust optimization
  (2023), \url{https://arxiv.org/abs/2301.11113}

\bibitem{mishra2021survey}
Mishra, S., Dutta, S., Long, J., Magazzeni, D.: A survey on the robustness of
  feature importance and counterfactual explanations. arXiv (v2) (2023),
  \url{https://arxiv.org/abs/2111.00358}

\bibitem{mothilal2020explaining}
Mothilal, R.K., Sharma, A., Tan, C.: Explaining machine learning classifiers
  through diverse counterfactual explanations. In: Proceedings of the
  conference on Fairness, Accountability, and Transparency (FAccT). pp.
  607--617 (2020)

\bibitem{mazzine2021framework}
de~Oliveira, R.M.B., Martens, D.: A framework and benchmarking study for
  counterfactual generating methods on tabular data. Applied Sciences
  \textbf{11}(16), ~7274 (2021)

\bibitem{pawelczyk2021carla}
Pawelczyk, M., Bielawski, S., van~den Heuvel, J., Richter, T., Kasneci, G.:
  {CARLA}: A python library to benchmark algorithmic recourse and
  counterfactual explanation algorithms. In: Conference on Neural Information
  Processing Systems (NeurIPS) -- Track on Datasets and Benchmarks. p.~17
  (2021)

\bibitem{PawelczykWWW20}
Pawelczyk, M., Broelemann, K., Kasneci, G.: Learning model-agnostic
  counterfactual explanations for tabular data. In: Proceedings of The Web
  Conference (WWW'20). pp. 3126--3132 (2020)

\bibitem{recourse_invalidation}
Pawelczyk, M., Datta, T., van-den Heuvel, J., Kasneci, G., Lakkaraju, H.:
  Probabilistically robust recourse: Navigating the trade-offs between costs
  and robustness in algorithmic recourse. In: Proceedings of the International
  Conference on Learning Representations (ICLR). OpenReview.net (2023)

\bibitem{poyiadzi2020face}
Poyiadzi, R., Sokol, K., Santos-Rodriguez, R., De~Bie, T., Flach, P.: Face:
  feasible and actionable counterfactual explanations. In: Proceedings of the
  AAAI/ACM Conference on AI, Ethics, and Society. pp. 344--350 (2020)

\bibitem{rawal2020algorithmic}
Rawal, K., Kamar, E., Lakkaraju, H.: Algorithmic recourse in the wild:
  Understanding the impact of data and model shifts. arXiv  \textbf{v3} (2020),
  \url{https://arxiv.org/abs/2012.11788}

\bibitem{upadhyay2021towards}
Upadhyay, S., Joshi, S., Lakkaraju, H.: Towards robust and reliable algorithmic
  recourse. Advances in Neural Information Processing Systems  \textbf{34},
  16926--16937 (2021)

\bibitem{ustun2019actionable}
Ustun, B., Spangher, A., Liu, Y.: Actionable recourse in linear classification.
  In: Proceedings of the conference on Fairness, Accountability, and
  Transparency (FAccT). pp. 10--19 (2019)

\bibitem{VanLooverenECML21}
Van~Looveren, A., Klaise, J.: Interpretable counterfactual explanations guided
  by prototypes. In: Proceedings of the European Conference on Machine Learning
  and Knowledge Discovery in Databases (ECML/PKDD). pp. 650--665 (2021)

\bibitem{VIRGOLIN2023103840}
Virgolin, M., Fracaros, S.: On the robustness of sparse counterfactual
  explanations to adverse perturbations. Artificial Intelligence  \textbf{316},
   103840 (2023)

\bibitem{Wachter2017CounterfactualEW}
Wachter, S., Mittelstadt, B.D., Russell, C.: Counterfactual explanations
  without opening the black box: Automated decisions and the {GDPR}. Harvard
  Journal of Law and Technology  \textbf{31}(2),  841--887 (2018)

\end{thebibliography}

\end{document}


\title{Generating robust counterfactual explanations\\--\\Supplementary material}

\author{
Victor Guyomard\inst{1,2}
\and Françoise Fessant\inst{1}
\and Thomas Guyet\inst{3}
\and Tassadit Bouadi\inst{2}
\and Alexandre Termier\inst{2}
}


\institute{
Orange Labs, Lannion, France\\
\email{victor.guyomard@orange.com}\\
 \and
Univ Rennes, Inria, CNRS, IRISA, Rennes, France \and
Inria, Centre de Lyon, France
}

\maketitle

\begin{abstract}
Counterfactual explanations have become a mainstay of the XAI field. This particularly intuitive statement allows the user to understand what small but necessary changes would have to be made to a given situation in order to change a model prediction. The quality of a counterfactual depends on several criteria: realism, actionability, validity, robustness, etc. In this paper, we are interested in the notion of robustness of a counterfactual. 
More precisely, we focus on robustness to counterfactual input changes. This form of robustness is particularly challenging as it involves a trade-off between the robustness of the counterfactual and the proximity with the example to explain. We propose a new framework, \CostRC, that generates robust counterfactuals while managing effectively this trade-off, and guarantees the user a minimal robustness. An empirical evaluation on tabular datasets confirms the relevance and effectiveness of our approach.
\keywords{Counterfactual explanation  \and Robustness \and Algorithmic recourse}
\end{abstract}

\section{Proofs}
\begin{lemma} \label{eq:lemma} We have for a given counterfactual $\cf{x}$:
\begin{equation}
\Gamma\left(\cf{x};p_{\varepsilon}\right) = \int_{\overline{V}} p_{\varepsilon} d\varepsilon
\end{equation}
where $\overline{V}=\{ \varepsilon \sim p_{\varepsilon} |  h\left( \cf{x}+\varepsilon\right) = 0\}$
\end{lemma}

\begin{proof}
The idea is to partition the integral on non-valid and valid examples: 
\begin{align*}
    \Gamma\left(\cf{x};p_{\varepsilon}\right)&= \E_{\varepsilon\sim p_{\varepsilon}}\left[1-h\left({\cf{x}}+\varepsilon\right)\right] \\
    &=\int_{\mathbb{R} ^{p}}1-h\left({\cf{x}}+\varepsilon\right) p_{\varepsilon} d\varepsilon \\
    &=\underbrace{\int_{\overline{V}}1-h\left({\cf{x}}+\varepsilon\right) p_{\varepsilon} d\varepsilon
    }_{\overline{V}=\{ \varepsilon \sim p_{\varepsilon} |  h\left( \cf{x}+\varepsilon\right) = 0\}} + 
    \underbrace{\int_{V}1-h\left({\cf{x}}+\varepsilon\right) p_{\varepsilon} d\varepsilon
    }_{V=\{ \varepsilon \sim p_{\varepsilon} |  h\left( \cf{x}+\varepsilon\right) = 1\}} \\
    &=\int_{\overline{V}} p_{\varepsilon} d\varepsilon
\end{align*}
\end{proof}

\begin{proposition}\label{res:upper_lower_our}%
Let $t\in[0,1]$ be a decision threshold and $\cf{x}$ be a counterfactual for an example $x\in\mathcal{X}$, an upper bound of the true recourse invalidation rate is given by:
\begin{equation}\label{eq:upper_lower_our}%
\Gamma\left(\cf{x};p_{\varepsilon}\right)\leq \frac{\Theta\left(\cf{x};p_{\varepsilon}\right) }{\left( 1-t\right) }
\end{equation}
\end{proposition}
\begin{proof}
    \begin{align*}
    \Theta\left(\cf{x};p_{\varepsilon}\right)&=\int_{\mathbb{R} ^{p}}1-f\left({\cf{x}}+\varepsilon\right) p_{\varepsilon} d\varepsilon \\
    &=\int_{V}1-f\left({\cf{x}}+\varepsilon\right) p_{\varepsilon} d\varepsilon + \int_{\overline{V}}1-f\left({\cf{x}}+\varepsilon\right)p_{\varepsilon} d\varepsilon \\
    &\geq \int_{\overline{V}}1-f\left({\cf{x}}+\varepsilon\right)p_{\varepsilon} d\varepsilon && \text{Because integrals of a positive function}
    \end{align*}
On non-valid examples we have:
\begin{align*}
    f(\cf{x} + \varepsilon) \leq t
\end{align*}
Then:
\begin{align*}
1-f(\cf{x} + \varepsilon) &\geq (1-t) \\
\int_{\overline{V}}1-f\left({\cf{x}}+\varepsilon\right)p_{\varepsilon} d\varepsilon &\geq (1-t) \int_{\overline{V}} p_{\varepsilon} d\varepsilon \\
&= (1-t) \Gamma\left(\cf{x};p_{\varepsilon}\right) && \text{By using Lemma \ref{eq:lemma}}
\end{align*}
\end{proof}

\begin{lemma} \label{eq:lemma_biased} $\tilde{\Theta}$ is a non-biased estimator:
\begin{equation*}
\E_{\varepsilon\sim p_{\varepsilon}}\left[\tilde{\Theta}\left(\cf{x};K,p_{\varepsilon}\right)\right]= \Theta\left(\cf{x};p_{\varepsilon}\right)
\end{equation*}

\end{lemma}
\begin{proof}
\begin{align*}
    \E_{\varepsilon\sim p_{\varepsilon}}\left[\tilde{\Theta}\left(\cf{x};K,p_{\varepsilon}\right)\right] &=\E_{\varepsilon\sim p_{\varepsilon}}\left[\frac{1}{K} \sum_{k=1}^K (1 - f(\cf{x} + \varepsilon_k))\right]  \\
    &= \frac{1}{K} \sum_{k=1}^K \E_{\varepsilon\sim p_{\varepsilon}}\left[1 - f(\cf{x} + \varepsilon)\right] && \text{Linearity of the expectation} \\
    &= \tilde{\Theta}\left(\cf{x};p_{\varepsilon}\right) && \text{Because $\varepsilon$'s are identically distributed}
\end{align*}

\end{proof}

\begin{proposition}\label{eq:our_estimator_quality}
Let $t\in[0,1]$ be a decision threshold, $p_{\varepsilon}$ a noise distribution, $\cf{x}$ be a counterfactual for an example $x\in\mathcal{X}$, then an almost-sure upper-bound of the recourse invalidation rate is given by:
\begin{equation}
\mathbb{P}\left( \Gamma\left(\cf{x};p_{\varepsilon}\right) \leq \dfrac{m+\tilde{\Theta}\left(\cf{x};K,p_{\varepsilon}\right)}{1-t}\right) \geq 1-\exp \left( -2m^{2}K\right) 
\end{equation}
where $m>0$ and $K$ is the number of random samples.
\end{proposition}
\begin{proof}
We have: $ \tilde{\Theta}\left(\cf{x};K,p_{\varepsilon}\right) = \frac{S\left(\cf{x};K,p_{\varepsilon}\right)}{K}$, where $S\left(\cf{x};K,p_{\varepsilon}\right)=X_{1}+X_{2}+\ldots +X_{K}$ and $X_{k}=\left(1 - f(\cf{x} + \varepsilon_k)\right)$.
For readability reasons, $S\left(\cf{x};K,p_{\varepsilon}\right)$ will be noted as $S_K$ in the proof. 
Since $f$ returns probabilities, we have: $0\leq X_{k}\leq 1$. Moreover $X_{1},\ldots ,X_{k}$ are independents random variables, then:
\begin{align*}
    \mathbb{P}\left(S_K-\E_{\varepsilon\sim p_{\varepsilon}}\left[S_k\right] \leq -mK \right) &= \mathbb{P}\left(\frac{S_K}{K}-\E_{\varepsilon\sim p_{\varepsilon}}\left[\frac{S_k}{K}\right] \leq -m \right) \\ 
    &= \mathbb{P}\left(\tilde{\Theta}\left(\cf{x};K,p_{\varepsilon}\right) - \Theta\left(\cf{x};p_{\varepsilon}\right) \leq -m \right) && \text{By using Lemma \ref{eq:lemma_biased}} \\ 
    &\leq \exp \left( -2m^{2}K\right) && \text{By using Hoeffding's inequality}
\end{align*}
Then:
\begin{align*}
\mathbb{P}\left(\Theta\left(\cf{x};p_{\varepsilon}\right) \leq m + \tilde{\Theta}\left(\cf{x};K,p_{\varepsilon}\right)  \right)
&= \mathbb{P}\left(\frac{\Theta\left(\cf{x};p_{\varepsilon}\right)}{(1-t)} \leq \frac{m+\tilde{\Theta}\left(\cf{x};K,p_{\varepsilon}\right)}{(1-t)} \right) && \text{Because $t<1$} \\
& \geq 1-\exp \left( -2m^{2}K\right)
\end{align*}
Finally by using proposition \ref{eq:upper_lower_our} we have: 
\begin{align*}
\mathbb{P}\left(\Gamma\left(\cf{x};p_{\varepsilon}\right) \leq \frac{m+\tilde{\Theta}\left(\cf{x};K,p_{\varepsilon}\right)}{(1-t)} \right) \geq 1-\exp \left( -2m^{2}K\right)
\end{align*}
\end{proof}

\section{On the choice of $m$ and $K$ parameters}

\begin{table}[]
\centering
\begin{tabular}{|l|l|}
\hline
(m,K)         & $1-exp(-2m^{2}k)$ \\ \hline
(0.1,100)     & 0.865                                \\ \hline
(0.1,250)     & 0.993                                \\ \hline
(0.1,500)     & 0.999                                \\ \hline
(0.01,100)    & 0.09                                 \\ \hline
(0.01,10 000) & 0.86                                 \\ \hline
(0.01,30 000) & 0.997                                \\ \hline
\end{tabular}
\caption{Confidence levels for different values of $m$ and $K$ regarding proposition \ref{eq:our_estimator_quality}}
\end{table}

\section{Complementary experiments}
\subsection{Validity metric evaluation:}

\begin{figure}[H]
    \centering
    \includegraphics[width=.8\textwidth]{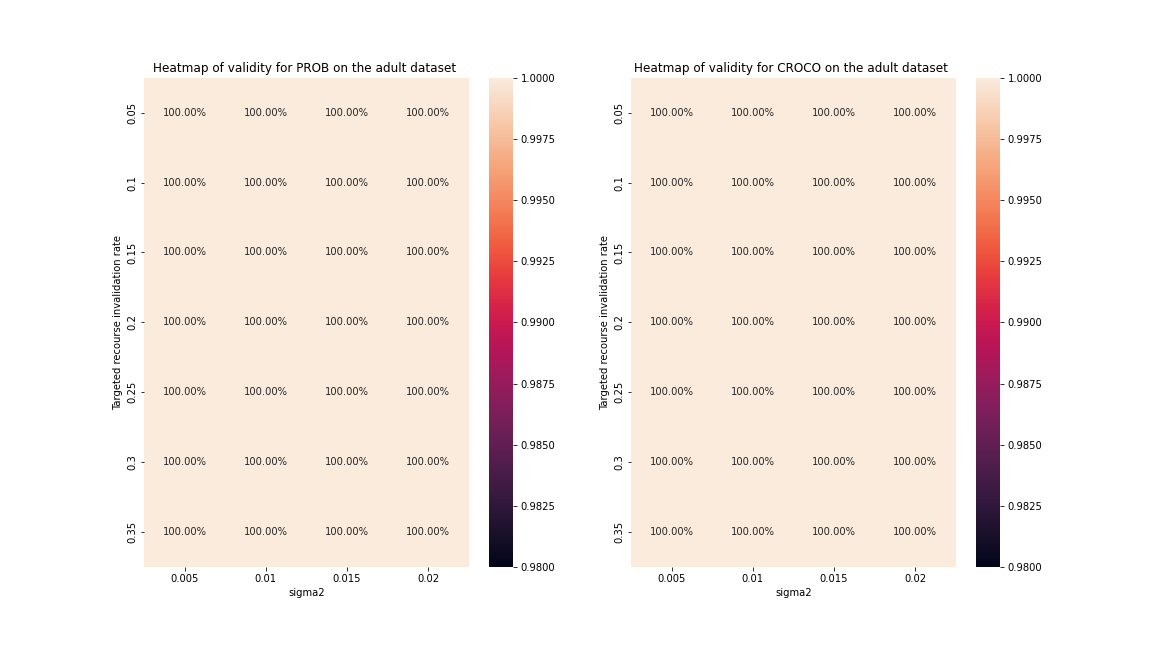}
    \caption{Heatmaps of validity metric for the adult dataset, with PROBE (right) and \CostRC (left). On the x-axis is values of $\sigma ^{2}\in \left\{ 0.005,0.01,0.015,0.02\right\}$ and on the y-axis values of $target\in \left\{0.05,0.10,0.15,0.2,0.25,0.3,0.35\right\}$. Each cell represents the percentage of validity for a given configuration. } 
    \label{fig:trade_off_adult}
\end{figure}

\begin{figure}[H]
    \centering
    \includegraphics[width=.8\textwidth]{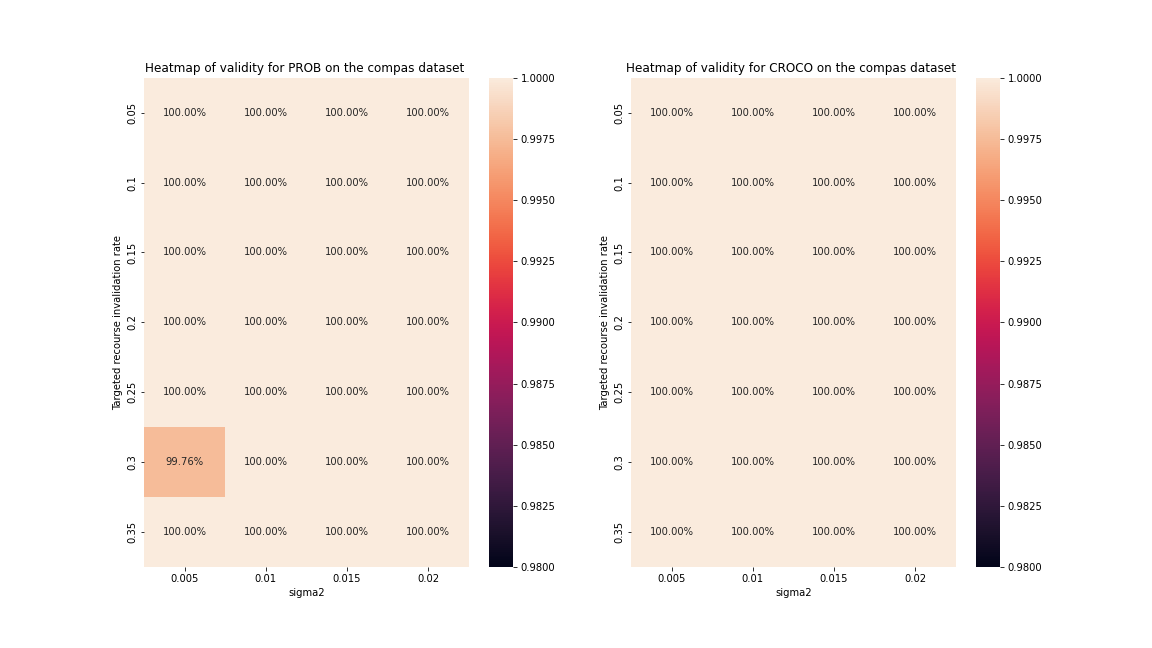}
    \caption{Heatmaps of validity metric for the compas dataset, with PROBE (right) and \CostRC (left). On the x-axis is values of $\sigma ^{2}\in \left\{ 0.005,0.01,0.015,0.02\right\}$ and on the y-axis values of $target\in \left\{0.05,0.10,0.15,0.2,0.25,0.3,0.35\right\}$. Each cell represents the percentage of validity for a given configuration.} 
    \label{fig:trade_off_adult}
\end{figure}

\begin{figure}[H]
    \centering
    \includegraphics[width=.8\textwidth]{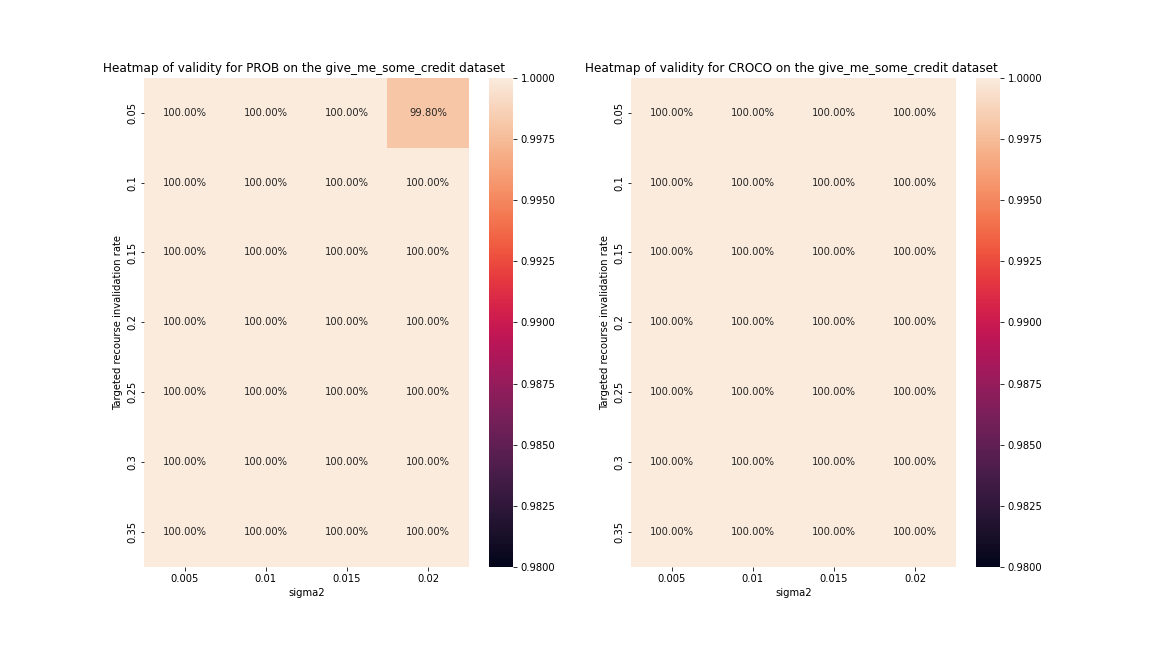}
    \caption{Heatmaps of validity metric for the GSC dataset, with PROBE (right) and \CostRC (left). On the x-axis is values of $\sigma ^{2}\in \left\{ 0.005,0.01,0.015,0.02\right\}$ and on the y-axis values of $target\in \left\{0.05,0.10,0.15,0.2,0.25,0.3,0.35\right\}$. Each cell represents the percentage of validity for a given configuration.} 
    \label{fig:trade_off_adult}
\end{figure}

\subsection{Standard-deviation values for PROBE and \CostRC}
\begin{figure}[H]
    \centering
    \includegraphics[width=.8\textwidth,trim={0 7pt 0 20pt},clip]{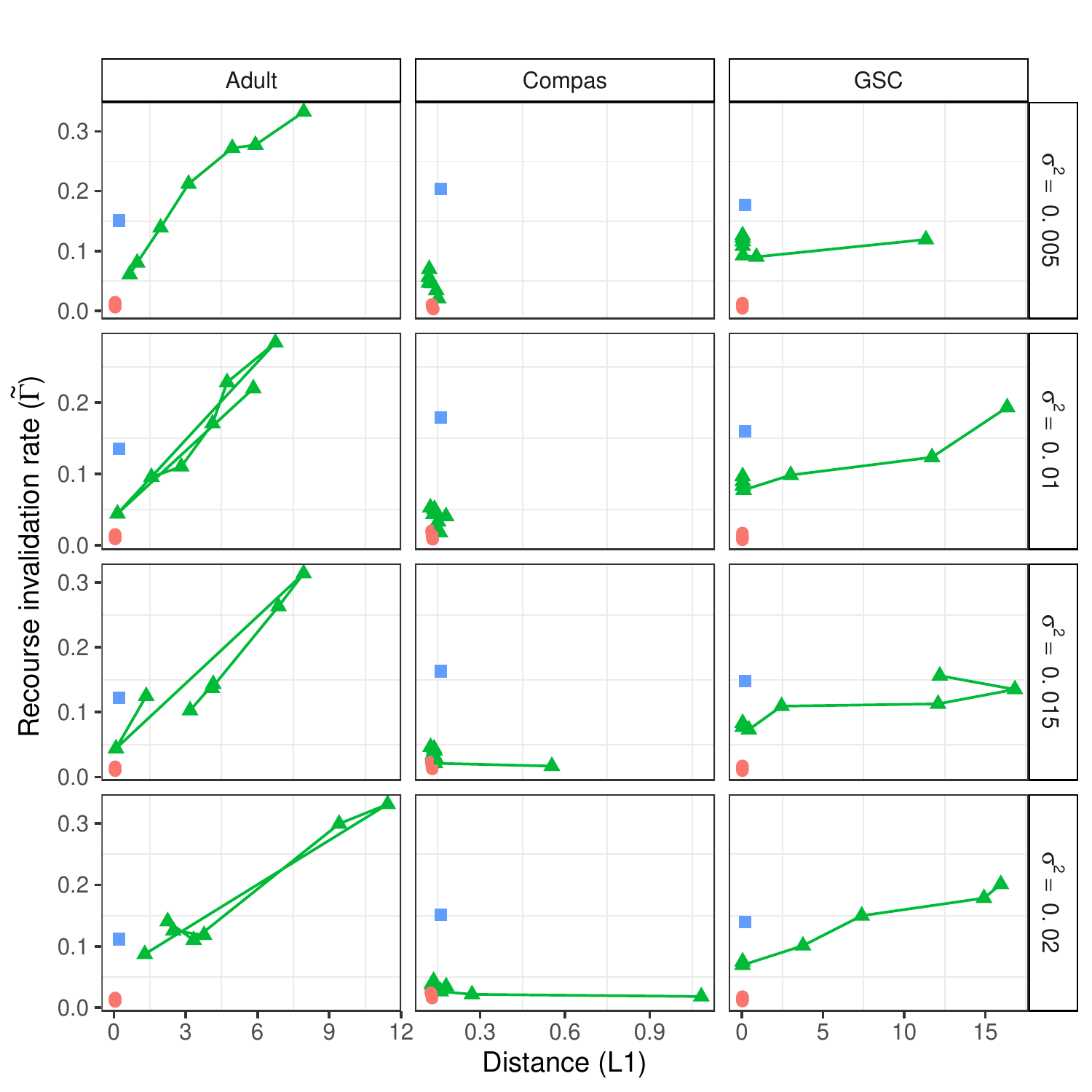}\\
    \includegraphics[width=.8\textwidth,trim={0 18.5pt 0 22pt},clip]{plots/legend_curves.pdf}
    \caption{Standard-deviation values regarding distance and recourse invalidation rate. Each column corresponds to a dataset and each line to a value of $\sigma^2 \in \left\{0.005,0.01,0.015,0.02\right\}$. In each subplot the value of  $\sigma ^{2}$ is fixed. Each point of a curve corresponds to the standard-deviations for recourse invalidation rate and distance, all for a given target, we have $target\in \left\{0.05,0.10,0.15,0.2,0.25,0.3,0.35\right\}$. The points are connected by target order.}
    \label{fig:trade_off_cost_IR}
\end{figure}

\subsection{Empirical validation of the upper-bound}
\begin{figure}[H]
    \centering
    \includegraphics[width=\textwidth,trim={0 6pt 0 20pt},clip]{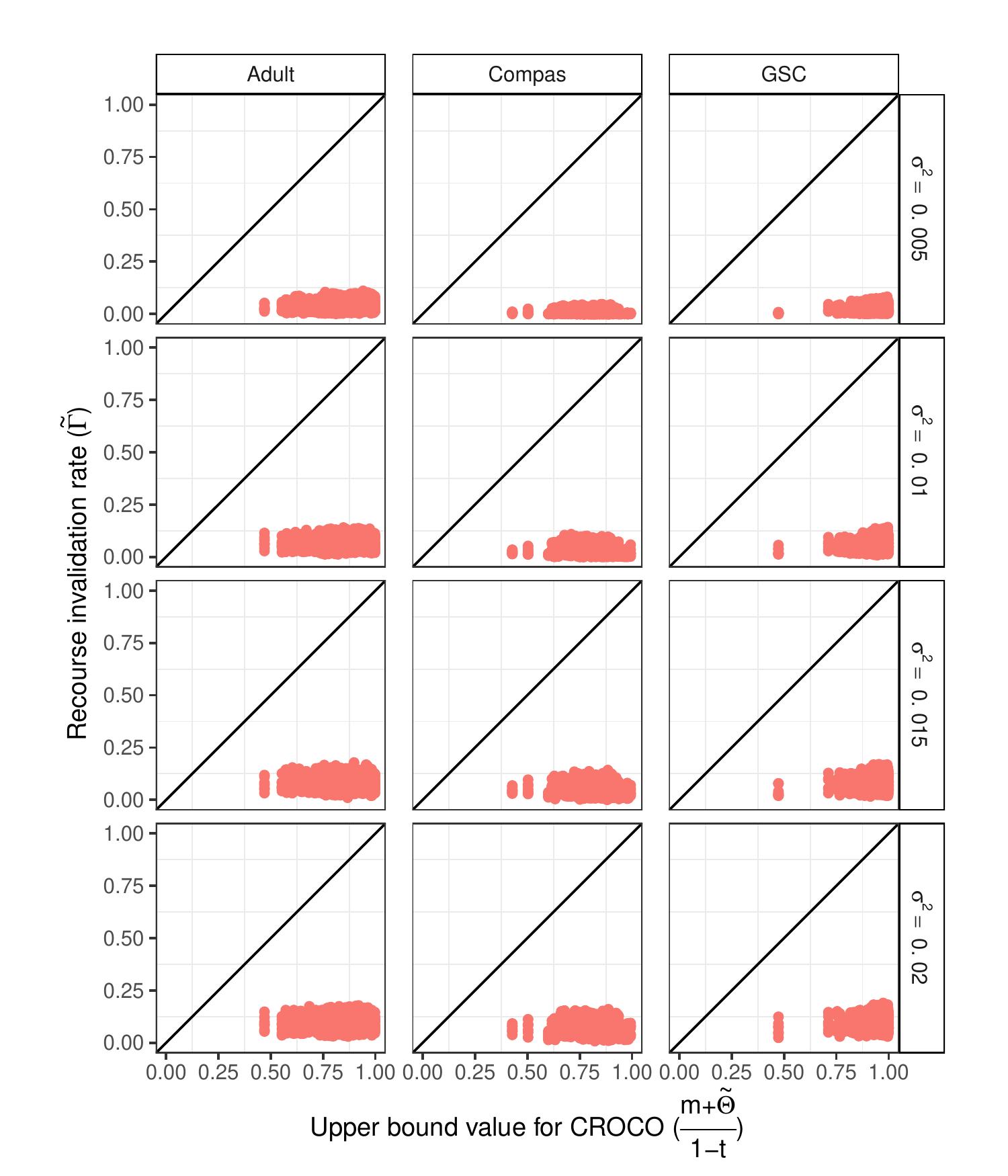}
    \caption{Empirical evaluation of the upper-bound. Each column corresponds to a dataset and each line to a value of $\sigma^2 \in \left\{0.005,0.01,0.015,0.02\right\}$. Each point represents a counterfactual. The x-axis corresponds to the upper-bound values for $m=0.1$ and $t=0.5$ computed after minimization, and the y-axis the computed recourse invalidation rate.  } 
    \label{fig:upper_bound_comparison}
\end{figure}


\newpage
\section{Dataset Details}
\subsection{Tabular Data} 
We used the CARLA library that contains 3 binary classification datasets. 
Table \ref{tab:dataset} describe the size of each dataset as well as the number of categorical and continuous variables. 
Each dataset is divided into a training and a testing set, the size are respectively $75\%$ and $25\%$.

\begin{table}[H]
\centering
\small
\caption{Datasets details}
\begin{tabular}{llcc}
\hline \text { Dataset } & \text { Size } & \text { Continuous } & \text { Categorical } \\
\hline \text { Adult } & 48,832 & 6 & 7 \\
\text { COMPAS } & 6172 & 4 & 3 \\
\text { GSC } & 115,527 & 10 & 0 \\
\hline
\label{tab:dataset}
\end{tabular}
\end{table}

\section{Implementation Details}

\subsection{\CostRC Code} 
The code to reproduce the results of the article is available at this link \url{https://github.com/twi09/CROCO}

\subsection{\CostRC Details} 
For each dataset we choose a learning rate $\alpha=0.001$.
The objective function $\mathcal{L}_{\CostRC}$ is minimized for different $\lambda$ values until the stopping criterion of algorithm \ref{algorithm:CostRC} is met. 
$\lambda$ is initialized to $1$ and decreased by $0.25$ at each step. 

\begin{algorithm}[t]
  \caption{\texttt{\CostRC} optimization for counterfactual generation}
  \label{algorithm:CostRC}
\begin{algorithmic}
   \State {\bfseries Input:} $x$ s.t.\ $f(x) < 0$, $f$, $\lambda > 0$, $\alpha$, $\bar\Gamma_{t}>0$, $K$,$p_\varepsilon$
   \State {\bfseries Output:} $x+\delta$
   \State $\delta\gets 0$;
   \State Compute $\tilde{\Theta}\left(x+\delta;K,p_\varepsilon\right)$
  \While{ $f(x+\delta)< 0$ {\bfseries and} $\frac{m+\tilde{\Theta}\left(x+\delta;K,p_\varepsilon\right)}{1-t}> \bar\Gamma_{t}$}
   \State $\delta \gets  \delta - \alpha \cdot \nabla_{\delta} \mathcal{L}_{\CostRC}(x+\delta; \Theta_{t}, p_\varepsilon, \lambda)$ 
   \State Update $\tilde{\Theta}\left(x+\delta;K,p_\varepsilon\right)$
  \EndWhile
\State {\bfseries Return:} $x+\delta$ 
\end{algorithmic}
\end{algorithm}

\subsection{Additional Information on PROBE Reproducibility}
We used the PROBE code that was available at this link \url{https://openreview.net/forum?id=sC-PmTsiTB}.
To ensure a fair comparison, we applied identical data preprocessing techniques and used the same machine learning models.